\pdfoutput=1

\documentclass[11pt]{article}

\usepackage{EMNLP2023}

\usepackage{times}
\usepackage{latexsym}

\usepackage{hyperref}

\usepackage[T1]{fontenc}

\usepackage[utf8]{inputenc}

\usepackage{microtype}

\usepackage{inconsolata}

\usepackage{graphicx}

\usepackage{covington}

\title{Crosslingual Structural Priming and the Pre-Training Dynamics of Bilingual Language Models}

\author{Catherine Arnett$^{a}$ ~~ Tyler A. Chang$^{b}$  ~~ James A. Michaelov$^{b}$~~  Benjamin K. Bergen$^{b}$\\
  $^{a}$Department of Linguistics, \\
  $^{b}$Department of Cognitive Science, \\
  University of California San Diego \\
  \texttt{\{ccarnett, tachang, j1michae, bkbergen\}@ucsd.edu}
  }
\begin{document}
\maketitle

\begin{abstract} 
Do multilingual language models share abstract grammatical representations across languages, and if so, when do these develop? Following \citet{sinclair_2022_StructuralPersistenceLanguage}, we use structural priming to test for abstract grammatical representations with causal effects on model outputs.
We extend the approach to a Dutch-English bilingual setting, and we evaluate a Dutch-English language model during pre-training. 
We find that crosslingual structural priming effects emerge early after exposure to the second language, with less than 1M tokens of data in that language. We discuss implications for data contamination, low-resource transfer, and how abstract grammatical representations emerge in multilingual models.
\end{abstract}

\section{Introduction}

Multilingual language models share representations across languages \citep{artetxe-etal-2020-cross, conneau-etal-2020-emerging}, which is thought to enable their crosslingual transfer abilities \citep{wu-dredze-2019-beto,Chi_Dong_Wei_Wang_Mao_Huang_2020,pmlr-v119-hu20b,winata-etal-2021-language,winata-etal-2022-cross}.
In this study, we use a paradigm known as crosslingual structural priming to study shared representations of grammatical structure in a Dutch-English bilingual language model.
Specifically, does crosslingual structural priming occur, and how much pre-training data does it require?

\textbf{Structural priming} is a phenomenon in which after being presented with a sentence with a given grammatical structure, people (and language models; \citealp{sinclair_2022_StructuralPersistenceLanguage}) are more likely to produce a sentence with the same structure \citep{bock_1986_SyntacticPersistenceLanguage,prasad-etal-2019-using,frank_2021_CrosslanguageStructuralPriming,li-etal-2022-neural,choi2022syntactic2}. For example, a language model would assign a higher probability to a prepositional object (PO) dative sentence (e.g. ``\textit{the chef gives a hat to the swimmer}'') following another PO sentence than it would following a double object (DO) dative sentence (e.g. ``\textit{the chef gives the swimmer a hat}''; sentences from \citealp{schoonbaert_2007_RepresentationLexicalSyntactic}).
Because the grammatical structure is primed rather than a specific semantic meaning, \citet{sinclair_2022_StructuralPersistenceLanguage} argue that structural priming effects provide evidence for abstract grammatical representations in language models.
By measuring output model probabilities given a prime sentence, structural priming demonstrates causal effects of grammatical representations on model outputs without relying on access to internal model states.
The presence of structural priming in crosslingual scenarios (e.g. a structure primes a similar structure in another language) would indicate that these representations are shared between languages.

\section{Method}

\subsection{Bilingual Model Pre-Training}
We pre-train a Dutch-English bilingual language model to simulate the language experience of the Dutch-English bilingual participants in \citet{schoonbaert_2007_RepresentationLexicalSyntactic}.
The model is an autoregressive GPT-2 Transformer language model with 124M parameters \citep{radford-etal-2018-improving,radford-etal-2019-language}.
The model is pre-trained on 6B tokens each of the Dutch and English OSCAR corpus \citep{AbadjiOrtizSuarezRomaryetal.2021} for 1M pre-training steps with batch size 128.
To simulate L1-L2 learning, the model is exposed only to Dutch data for the first half of pre-training. During the second half of pre-training, the model is given an equal mix of Dutch and English data.
The tokenizer is trained on 25\% English text and 75\% Dutch text to match the proportions of data from each language seen during pre-training.

\subsection{Materials}
We test our bilingual model for crosslingual structural priming using the stimuli from \citet{schoonbaert_2007_RepresentationLexicalSyntactic}.
Structural priming studies measure how frequently speakers produce sentences with different grammatical structures after corresponding prime structures.
Our study uses the \textbf{dative alternation}, where sentences can either be expressed with a PO or DO construction (see Introduction).
We consider Dutch primes with English targets.

\subsection{Calculating Structural Priming}
Following human studies \citep{loebell_2003_StructuralPrimingLanguages,schoonbaert_2007_RepresentationLexicalSyntactic}, we compute the normalized probability of each target sentence following each prime. For example, we compute the normalized probability $P_N$ of a PO target $T_{PO}$ following a PO prime $P_{PO}$ as shown below, where $T_{DO}$ is the DO target and $P_{DO}$ would be a DO prime:
$$P_N(T_{PO}|P_{PO}) = \frac{P(T_{PO}|P_{PO})}{P(T_{PO}|P_{PO})+P(T_{DO}|P_{PO})}$$
To test for a structural priming effect, we compare $P_N(T_{PO}|P_{PO})$ and $P_N(T_{PO}|P_{DO})$. If the former is significantly higher, this would indicate structural priming, because PO targets are more likely after PO primes than after DO primes.

\section{Results}
\begin{figure}[t]
    \centering
    \includegraphics[width=\columnwidth]{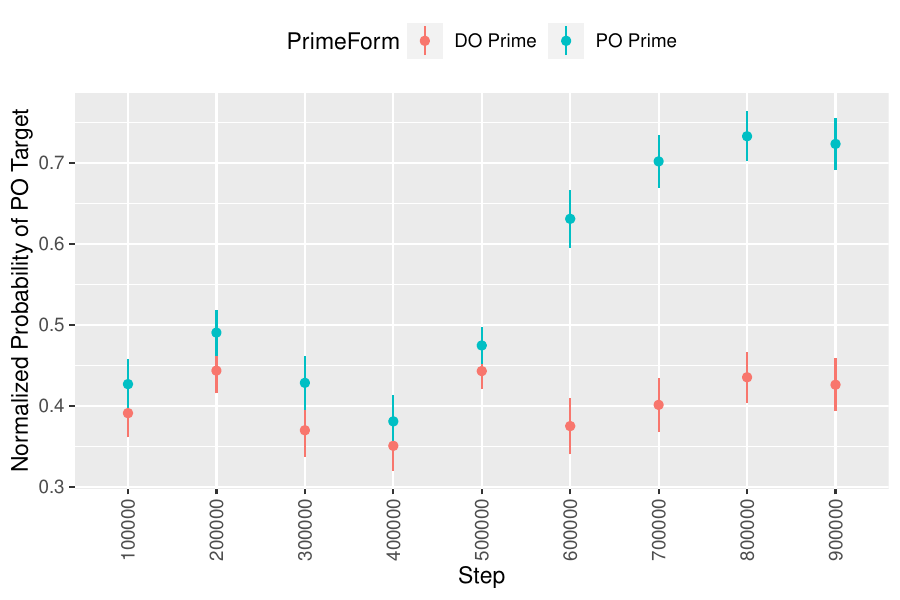}
    \caption{Language model priming effect for Dutch-English structural priming experiments every 100K steps during pre-training. The model is first exposed to English at 500K steps.}
    \label{fig:xlang_effect}
\end{figure}

As shown in \autoref{fig:xlang_effect}, the English PO target probabilities are numerically higher after PO primes than after DO primes throughout pre-training, but the effect drastically increases after the model begins training on English.\footnote{We hypothesize that apparent crosslingual priming effects before 500K steps may be due to English contamination in the Dutch pre-training data.}
To quantify the point at which the structural priming effect emerges, we consider results for the first 200 steps after the model is first exposed to English, in 10-step intervals.
We fit a linear mixed-effects model predicting normalized PO target probability based on prime type and pre-training step.\footnote{We treat pre-training step as a categorical rather than continuous variable, because effects of pre-training step may be nonlinear. We include a random intercept for stimulus item. Structural priming at step $t$ (an effect of prime type at step $t$ beyond the effect at step 500K) is reflected as an interaction term between pre-training step $t$ and prime type.}
We find a significant overall interaction between prime type and pre-training step ($\chi^2(20)=56.86, p<0.001$), suggesting that the magnitude of the structural priming effect changes throughout pre-training.
The effect of prime type is first significantly different from step 500K (after correcting for multiple comparisons) at step 500120, or 120 steps after English training begins ($t(3943)=2.93$, adjusted $p=0.030$).
At that point, the model has been exposed to 983040 English tokens.

\setlength{\belowcaptionskip}{-1cm}

\section{Discussion}

We find evidence of structural priming after 120 steps of English training, suggesting that crosslingual grammatical representations can emerge with fewer than 1M tokens of data in a secondary language.
This result is important from multiple perspectives. First, understanding how much data is needed to obtain shared multilingual representations has implications for transfer learning to low-resource languages (e.g., \citealp{winata-etal-2022-cross,ogueji-etal-2021-small}). 
Our results suggest that structures in a high-resource language can quickly transfer to a new language, although our results only consider a pair of highly related languages, Dutch and English.

Second, these results illuminate the effects of crosslingual data contamination.
Ongoing research has demonstrated that contamination with other languages can impact multilingual model performance \citep{blevins-zettlemoyer-2022-language,muennighoff-etal-2023-crosslingual}. \citet{muennighoff-etal-2023-crosslingual} find zero-shot crosslingual transfer on XNLI for models that are not intentionally trained on some of the XNLI languages. In an analysis of the pre-training dataset, they find small amounts of data in non-included languages (e.g. approximately 0.006\% of the data is in Thai, corresponding to roughly 20M tokens).
Our work demonstrates that it is possible to observe crosslingual effects with fewer than 1M tokens in a target language.
This is confirmatory evidence that data contamination may be driving apparent ``zero-shot'' crosslingual capabilities in multilingual language models.

\section*{Acknowledgements}
We would like to thank Tiffany Wu, Fiona Tang, Emily Xu, and Jason Tran for helping to prepare stimuli.
Models were pre-trained and evaluated using hardware provided by the NVIDIA Corporation as part of an NVIDIA Academic Hardware Grant.
Tyler Chang is partially supported by the UCSD HDSI graduate fellowship.

\bibliography{library,custom,custom2}
\bibliographystyle{acl_natbib}

\end{document}